\documentclass[sigconf]{acmart}
\usepackage{popets}
\usepackage{xspace}
\usepackage{microtype}
\usepackage{graphicx}
\usepackage{subfigure}
\usepackage{booktabs} 
\usepackage{graphicx,subfigure}

\setcopyright{popets}
\copyrightyear{}

\acmYear{2023}
\acmVolume{XX}
\acmNumber{XX}
\acmISBN{XX}
\acmConference{Conference '23}
\acmDOI{XXXXXXX.XXXXXXX}
\begin{document}

\title[]{P4: Towards private, personalized, and Peer-to-Peer learning}


\author{Mohammad M Maheri}
\affiliation{%
  \institution{Imperial College London}
  \city{London}
  \state{}
  \country{UK}}
\email{m.maheri23@imperial.ac.uk}

\author{Sandra Siby}
\affiliation{%
  \institution{Imperial College London}
  \city{London}
  \state{}
  \country{UK}}
\email{s.siby@imperial.ac.uk}

\author{Sina Abdollahi}
\affiliation{%
  \institution{Imperial College London}
  \city{London}
  \state{}
  \country{UK}}
\email{s.abdollahi22@imperial.ac.uk}

\author{Anastasia Borovykh}
\affiliation{%
  \institution{Imperial College London}
  \city{London}
  \state{}
  \country{UK}}
\email{a.borovykh@imperial.ac.uk}

\author{Hamed Haddadi}
\affiliation{%
  \institution{Imperial College London}
  \city{London}
  \state{}
  \country{UK}}
\email{h.haddadi@imperial.ac.uk}


\newcommand{\para}[1]{\smallskip \noindent \textbf{#1}}
\newcommand{\parait}[1]{\smallskip \noindent \textit{#1}}

\newcommand\eg{\emph{e.g.},\xspace}
\newcommand\ie{\emph{i.e.},\xspace}
\newcommand\etc{\emph{etc}.\xspace}
\newcommand\via{\emph{via}}
\newcommand{\name}{\textsc{P4}\xspace}
\providecommand{\etal}{\emph{et al.}\xspace}

\begin{abstract}
Personalized learning is a proposed approach to address the problem of data heterogeneity in collaborative machine learning.
In a decentralized setting, the two main challenges of personalization are client clustering and data privacy. 
In this paper, we address these challenges by developing \name (Personalized Private Peer-to-Peer) -- a method that ensures that each client receives a personalized model while maintaining  differential privacy guarantee of each client's local dataset during and after the training. 
Our approach includes the design of a lightweight algorithm to identify similar clients and group them in a private, peer-to-peer (P2P) manner. 
Once grouped, we develop differentially-private knowledge distillation for clients to co-train with minimal impact on accuracy. 
We evaluate our proposed method on three benchmark datasets (FEMNIST or Federated EMNIST, CIFAR-10 and CIFAR-100) and two different neural network architectures (Linear and CNN-based networks) across a range of privacy parameters. The results demonstrate the potential of \name, as it outperforms the state-of-the-art of differential private P2P by up to $40\%$ in terms of accuracy. 
We also show the practicality of \name by implementing it on resource-constrained devices, and validating that it has minimal overhead, \eg $\approx$ 7 seconds to run collaborative training between two clients. 
\end{abstract}

\keywords{Peer-to-Peer machine learning, decentralized learning, personalization, differential privacy}

\maketitle

\section{Introduction}
\label{sec:introduction}
Collaborative learning involves training a global machine learning model by sharing local models across clients without compromising data privacy.  
However, client data heterogeneity -- differences in data distribution or tasks across clients -- means that a global model may not satisfy all clients. 
Recently, there has been growing interest in overcoming data heterogeneity issues by considering personalization for each client \cite{li2022learning,collins2021exploiting,li2021ditto}. 
\textit{Personalized learning} attempts to adapt each client model to its specific task or distribution while effectively leveraging shared knowledge among clients. 
A decentralized or peer-to-peer (P2P) setting, where clients have their own distributions but can benefit from sharing information with \textit{some} clients, has been particularly considered as intuitive for personalization~\cite{zantedeschi2020fully, li2022learning}.

There are two main challenges in personalized decentralized learning.
The first challenge is \textit{clustering}, where
clients attempt to find similar clients to share knowledge with. 
Clustering algorithms \cite{li2022towards,duan2021flexible} have been used to group clients with similar data distributions and train a global model on each group. 
It has been shown that this could improve the accuracy of the trained model. 
However, most of the proposed methods depend on the number of clusters and need a central server to conduct clustering \cite{xie2021multi,sattler2020clustered,nguyen2022self,briggs2020federated}. 
This is not a practical assumption in most applications and comes with communication bandwidth issues. 
More importantly, these works do not consider privacy leakages that may arise as a result of clients sharing data in order to perform clustering. 
The second challenge is \textit{data privacy} -- client data that is used to train models may contain sensitive information. 
Therefore, the training algorithm must provide privacy guarantees regarding clients’ data. 
A popular approach to address this challenge is by computing the sensitivity of clients’ local data and adding differential privacy (DP) noise and clipping gradients~\cite{li2019asynchronous, wei2020federated, truex2020ldp, geyer2017differentially}. 
At the same time, adding random noise and clipping gradients results in degrading the performance of the trained model~\cite{bagdasaryan2019differential,noble2022differentially,wei2020federated}. 
In a decentralized setting, the effect of random noise on model accuracy is more severe due to the smaller number of model updates (gradients) that are used in model aggregation~\cite{kalra2023decentralized}. 

In this paper, we propose \name (Personalized Private Peer-to-Peer), a method to provide a personalized model for each client under differential privacy guarantee. 
\name addresses both the heterogeneity and the privacy challenges of DL.
In \name, clients first group themselves in a fully decentralized manner using a similarity metric based on their model weights. 
Once grouped, clients perform collaborative training with upper-bounded data privacy risk. 
\name uses handcrafted features to enhance the privacy-utility trade-off, complemented by a proxy model \cite{zhang2018deep}, to facilitate knowledge distillation among clients.
This separation between the proxy model and private model results in fast convergence of the learning process and personalized parameters for each client (as shown in Section \ref{sec:evaluation}). 
\name is applicable even to devices that have limited computation power, a prevalent characteristic in many real-world scenarios \cite{dhar2021survey,cui2018survey}.
It can achieve test accuracy improvement of ~6\% to ~40\
over state-of-the-art works under different levels of privacy boundaries ($\epsilon \in [3,20]$) using one linear layer and limited communication overhead (100 training iterations)

Our main contributions are as follows:

\begin{itemize}
    \item We develop \name, an approach to address the challenges of clustering and privacy in P2P learning. Our approach effectively addresses personalized knowledge sharing challenges while considering the limited computation and communication power of resource-constrained devices. It does so by collaborative training among clients that have similar data distribution and a new architecture that enables knowledge distillation between the proxy model and the private model.

    \item We propose a lightweight procedure for honest-but-curious clients to group themselves to perform collaborative training in a decentralized setting. In our procedure, clients group themselves using the l1-norm between their model weights as a similarity metric. We find that grouping based on this metric produces $\approx$ 10\% improvement in accuracy over random grouping.

    \item We introduce a method for clients to collaboratively train while maintaining privacy of their data. We find that knowledge distillation by proxy models can be improved in terms of accuracy and communication cost if we use  handcrafted features that are more robust to DP noise. This results in better accuracy-privacy trade-off by approximately ~7\% to ~9\% even with one shallow neural network. 
    
    \item We implement \name on Raspberry Pis and find that it has minimal overhead in terms of run time, memory usage, power consumption, and communication bandwidth. Our experiments show that \name is practical enough to be run on resource-constrained devices. 
    
\end{itemize}

To the best of our knowledge, our paper is the first to develop and study learning personalized models under differential privacy for fully decentralized (P2P) deep learning tasks.  
The code can be accessed at \href{https://anonymous.4open.science/r/private_decentralized_learning-F413/README.md}{this link}.

\section{Background and related work}
\label{sec:background}
We first review previous works that addressed the problems of privacy in decentralized learning and proposed methods to make learning algorithms personalized. We then discuss previous works that aimed to eliminate reliance on a centralized server or third-party in P2P learning.

\subsection{Private decentralized learning}
\label{subsection:private_fl}
In decentralized learning, the preservation of both data privacy and local model privacy for each client is crucial. 
Local model privacy is considered valuable because it has been shown that deep neural networks can inadvertently memorize training data, making them susceptible to model inversion attacks~\cite{geiping2020inverting, huang2021evaluating}. 
To address these concerns, various privacy-preserving methods have been proposed to safeguard plaintext gradients. 
Differential privacy has been employed by some studies to mitigate the privacy issues in decentralized learning~\cite{li2019asynchronous, wei2020federated, truex2020ldp, geyer2017differentially}. 
However, the introduction of noise to the gradients generated by each user as part of the differential privacy mechanism can lead to a degradation in model performance~\cite{truex2019hybrid}.  
Alternatively, certain approaches utilized homomorphic-encryption-based methods to preserve client privacy by encrypting and decrypting model weights or gradients~\cite{aono2017privacy, zhang2020privacy,zhao2022pvd}. 
However, they incur computation and communication overhead~\cite{jin2023fedml}, making them less suitable for edge devices with limited computational power. 
Other work used secure multi-party computation (MPC)~\cite{bonawitz2017practical, mandal2018nike, zheng2019helen}. They ensure privacy of data and model in federated learning by enabling collaborative computation among multiple parties while keeping their individual inputs confidential. 
However, these approaches have high communication cost. 
Recently, some works used trusted execution environments (TEE) to protect privacy~\cite{mo2021ppfl, mo2022sok, dhasade2022tee}. 
The main disadvantage of using hardware-based methods such as TEE is that all users may not have access to such hardware and they may have computational overhead.

\subsection{Personalized decentralized learning}
\label{subsection:personalized_fl}
The presence of non-iid client data necessitates the pursuit of personalization. 
The aim is to achieve customized individual models for each client through decentralized learning, which is a departure from a singular consensus global model. 
Li \etal~\cite{li2021ditto} leveraged regularization to encourage the personalized models to be close to the optimal global model. 
Zhang \etal~\cite{zhang2020personalized} calculated the optimal weighted combination of all client models as a personalized update for each client. 
There is another line of work that tries to cluster clients participating in decentralized learning to achieve better performance~\cite{ghosh2020efficient,sattler2020clustered}. 
In these works, aggregation of models is conducted in each cluster separately as combining models within a cluster improves personalization, but merging models from different clusters may lead to negative transfer \cite{li2022towards}. 
The primary obstacle in federated learning with clustering lies in assessing client similarity. There are primarily three types of measurement methods, based on losses~\cite{ghosh2020efficient,li2022towards}, gradients~\cite{sattler2020clustered,duan2021flexible}, and model weights \cite{xie2021multi,sattler2020clustered,nguyen2022self,briggs2020federated}, respectively. 
Except Li \etal~\cite{li2022towards}, all other approaches depend on a centralized server, and hence are not applicable in a P2P scenario. Moreover, most mentioned work, including Li \etal~\cite{li2022towards}, did not consider the data privacy leakage of clients when computing similarity. 
On the other hand, in loss-based methods, each client needs to receive other clients' data and compute the loss of its data on the received model to compute its similarity with other clients. 
This procedure involves receiving all models from all clients, which has a significant communication cost and computing the loss of the local data on all of them has a significant computation overhead. 
Moreover, most of them rely on the assumption of prior-knowledge of the number of clusters between clients, so clustering performance would degrade if the number of clusters was set inappropriately.

\subsection{P2P learning}
\label{subsection:decentralized_fl}

Fully decentralized learning (peer-to-peer learning) has emerged as a promising solution to address the challenges associated with centralized federated learning and personalization by eliminating the need for a central server and distributing computation among participating clients. 
This approach leverages direct communication between clients, offering potential benefits such as faster convergence~\cite{sun2022decentralized, li2022learning} and enhanced data privacy compared to centralized federated learning~\cite{shi2023improving, kairouz2021advances}. 
In this approach, clients communicate with a limited number of neighbors depending on their communication and computational power~\cite{lalitha2019peer,warnat2021swarm}.

Several studies have explored decentralized federated learning and proposed various techniques to optimize its performance. Some works employed a Bayesian approach to model shared knowledge among clients in a decentralized setting~\cite{lalitha2018fully, lalitha2019peer}. 
Roy \etal~\cite{roy2019braintorrent} introduced a server-less federated learning approach specifically designed for dynamic environments. 
Sun \etal~\cite{sun2022decentralized} focused on reducing communication and computation costs through quantization techniques, while Dai \etal~\cite{dai2022dispfl} proposed sparse training methods to achieve similar goals.

Several approaches have been proposed to address the challenge of heterogeneous data distributions in peer-to-peer learning. Dai \etal~\cite{dai2022dispfl} utilized personalized sparse masks to tailor the models of individual clients. Shi \etal~\cite{shi2023towards} introduced a personalized federated learning framework that incorporates decentralized partial model training using the Sharpness Aware Minimization (SAM) optimizer~\cite{foret2020sharpness} to mitigate model inconsistency. They further enhanced their approach by combining SAM with Multiple Gossip Steps (MGS)~\cite{ye2020decentralized} to overcome both model inconsistency and overfitting~\cite{shi2023improving}. 
Jeong \etal~\cite{jeong2023personalized} tackled the challenge of heterogeneous data distributions by employing weighted aggregation of model parameters based on the Wasserstein distance between the output logits of neighboring clients.

However, most existing works in peer-to-peer learning adopt random or predefined methods to select communication links between clients, without considering whether these connections would be beneficial for collaborative learning~\cite{shi2023towards, dai2022dispfl, shi2023improving, kalra2023decentralized, wang2023enhancing}. 
While Jeong \etal~\cite{jeong2023personalized} attempted to incorporate a dynamic topology based on client similarity, their approach solely relies on considering the similarity of output logits from clients' models, without taking into account the underlying distribution of their training data.
This limitation highlights the need for more intelligent and adaptive approaches to establish communication links between clients, with the goal of improving the overall learning process and faster convergence in presence of heterogeneous data distribution among clients.

Furthermore, while differential privacy has been explored in centralized federated learning \cite{noble2022differentially}, a significant gap exists in addressing data privacy and model privacy concerns in peer-to-peer learning approaches~\cite{shi2023towards, wang2022accelerating, dai2022dispfl, jeong2023personalized, shi2023improving}. Protecting the privacy of client data 
should be paramount to maintaining trust and security within the decentralized learning framework. 
Some approaches have been proposed to enhance privacy in peer-to-peer learning. Firstly, Bellet \etal~\cite{bellet2018personalized} studied the trade-off between utility and privacy in peer-to-peer FL. Moreover, Kalra \etal~\cite{kalra2023decentralized} introduced proxy models to facilitate efficient information exchange without a central server, leveraging differential privacy analysis to enhance privacy. However, this approach may introduce performance degradation caused by added differential privacy noise especially when the data is distributed non-IID among clients. Moreover, as mentioned before, merging models from different underlying data distributions would have a negative effect on performance. Most recently, Bayrooti \etal~\cite{bayrooti2023differentially} generalized DP-SGD to decentralized learning, aiming achieve consensus on model parameters for all clients after training. The consensus model may not perform well in scenarios where clients' tasks are non-IID.
Another approach, presented by Wang \etal~\cite{wang2023enhancing}, utilized gradient encryption algorithms to protect data privacy and employed succinct proofs to verify gradient correctness. However, this approach introduced computational and communication overheads while neglecting the heterogeneity of data distribution. Therefore, future research should emphasize the development of privacy-preserving mechanisms to safeguard sensitive information during the collaborative learning process.

\section{Methodology}
\label{sec:methodology}
In this section, we first formulate the problem in more detail, then we describe \name, our approach to enable clients to privately and efficiently co-train with other clients in a decentralized system.

\subsection{Problem formulation}

We consider $M$ clients $\{c_i\}_{i=1}^M$. Each client, $c_i$, holds a local dataset including data points and corresponding labels $\{X_i,Y_i\} \sim D_i$ drawn from its personal distribution $D_i$.
Each $c_i$ aims to learn personalized parameters $w_i$ to minimize the expected loss over the client's data distribution: 
\begin{equation}
    \label{eq:loss_minimization}
        F_i(w_i) = \mathbb{E}[ \mathcal{L}_{(x,y) \sim D_i} (w_i ; (x,y)) ]
\end{equation}

To find parameter to minimize the expected loss, each client seeks to determine an optimal parameter set, denoted as $w_i^*$, within the parameter space $w_i \in \mathbb{R}^d$. 
This optimization aims to minimize the expected loss over the client's own local data distribution $\ell (f_i,y)$, represented in Equation \ref{eq:w_star}: 

\begin{equation}
    \label{eq:w_star}
    w_i^* = arg \min_{w_i \in \mathbb{R}^d}  \mathbb{E}_{(x,y)\sim D_i} \ell (f_i (w_i ; x) , y)]
\end{equation}

However, solely using a client's local dataset is not sufficient to find a generalizable parameter for each client, so local training leads to poor generalization performance.
In order to achieve good generalization, each client is willing to collaborate and share knowledge (\ie their model gradients) with a \emph{subset} of other clients.
We note that the goal of the learning is not to reach a `global consensus' model, but for each client to obtain personalized models that perform well on their own data distributions. 

\para{Challenges.} There are three main challenges in our scenario. First, there is no central server that has strong computational power and is trusted by all clients. Hence, knowledge sharing should be decentralized among all the clients. 
Second, clients want to protect their data privacy, so the knowledge that is shared between clients should not violate the differential privacy of the clients’ data. Following previous works, this can be done by adding Gaussian noise to the shared gradients, but comes at the cost of performance degradation. Furthermore, in a fully decentralized setting, the number of clients that can participate in the collaborative training and aggregate their knowledge should be smaller than in centralized federated learning. Hence, the noise effects could have more devastating effects on the performance, as shown in the experiment section \ref{section:tradeoffDP}.
Third, each client has limited computation and communication power. 
Therefore, the learning algorithm for knowledge sharing should efficiently converge in a limited number of iterations, and the knowledge sharing method should be designed to minimize communication and computation overhead.

\subsection{Threat model}

We consider the honest-but-curious setting, where clients do not deviate from the training protocol and do not generate adversarial results, but are curious about the information of the other clients that are  involved in training. 
We consider a fully decentralized setting where there is no trusted third-party or server in the system with sufficient communication and computational power to handle the aggregation of all the clients in the network. 
We assume that clients communicate with one another over an encrypted channel and that a third-party adversary cannot eavesdrop on or modify this communication.
In our scenario, clients do not share their data directly with other clients in the network.
To ensure differential privacy guarantee for each client’s data, we prevent clients from resuming co-training once their privacy budget is exhausted.

\subsection{Design}
\label{section:design}
We propose \name -- a fully decentralized learning approach that facilitates knowledge sharing among clients with similar data distributions, by aggregating parameter updates from neighboring models. 
\name alternates between local updates and model aggregation, aiming to achieve personalized models for each client while minimizing computational and communication overheads. 

\begin{figure}[t!]
    \centering
    \includegraphics[width=1\linewidth]{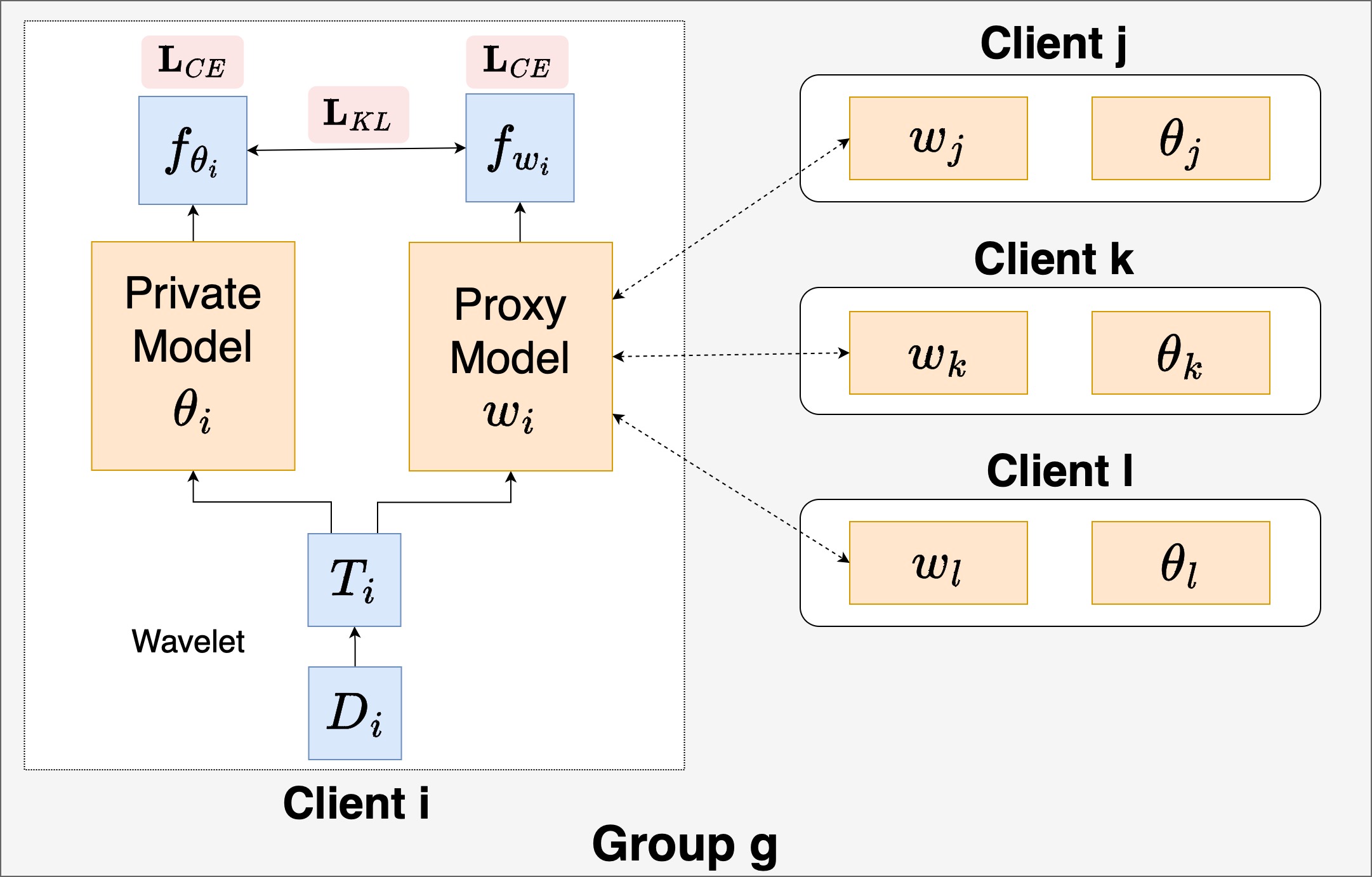}
    \caption{\textbf{Visualization of Model Aggregation in Group $g$:} Clients employ proxy and private models, aggregating only the proxy model. In each group, updates are exchanged with one client, serving as an aggregator, which can change during training to distribute communication overhead. After receiving the aggregated model, clients perform local training for personalization.}
    \label{fig:overview}
\end{figure}

\name consists of two phases: group formation and private co-training. 
In the group formation phase, clients search to identify peers with similar data distributions, forming client groups based on these similarities. 
Subsequently, in the private co-training phase, clients within a group share their locally computed data gradients with added noise, exclusively with other group members. 
Each client then aggregates the gradients it received from neighboring clients and updates its model.
Figure~\ref{fig:overview} illustrates the methodology.
We describe the phases in more detail below.

\para{Phase 1 Group formation:}
\label{subsection:similar_clients} 
The selection of clients without considering the underlying data distribution may lead to decreased performance compared to local training \cite{zec2023private}. 
Hence, in the first phase (\ie group formation), the objective of a client is to identify other clients with similar underlying data distributions to co-train with. 
This approach results in the partitioning of clients into distinct groups.
Instead of co-training with all clients, as typically done in prior research, each client co-trains only with the members of its group, which significantly reduces the communication and computational overhead (our measurements in Section~\ref{subsetion:energy} show that group formation has minimal overhead compared to training).
while improving the model accuracy of clients by providing better personalized models, as shown in Section \ref{sec:ablation}.

In \name, we define the client dissimilarity metric as the l1-norm between their model weights shown in Equation \ref{eq:l1_similarity}.  Inspired by \cite{ghosh2020efficient} and \cite{sattler2020clustered}, we utilize model weight similarity as an approximation of data distribution similarity among clients. Their research has shown that aggregating clients with similar gradients can lead to improved performance in centralized federated learning, and we extend this concept to the peer-to-peer (P2P) setting.

\begin{equation}
\label{eq:l1_similarity}
\text{dissimilarity}(i,j) = \lVert \text{vec}(w_i) - \text{vec}(w_j) \rVert_1,
\end{equation}
where $vec(w_i)$ is the flattened version of $w_i$.

Clients segregate themselves into groups based on the dissimilarity metric.
If we assume $M$ as the number of all clients in the network, this segregation can be depicted using a binary matrix $G \in {0,1}^{M \times M}$, where an entry $G_{ij}$ is one if clients $i$ and $j$ are part of the same group, and zero otherwise. The objective function is formulated as follows:

    \begin{equation}
        \underset{G}{min} \sum_{i=1}^M \sum_{j=1}^M  \mathbb{I}[G(i,j)=1] \lVert \text{vec}(w_i) - \text{vec}(w_j) \rVert_1
    \end{equation}

    such that:
    \begin{equation}
        \sum_{j=1}^{M} G_{i,j} \leq T \quad \text{for } i = 1,2,\ldots,M
    \end{equation}

where $G$ is the collaboration graph (symmetric matrix) and 
$T$ defines maximum group size (due to limited computation of each client). We note that \name enforces mutual collaborative training, implying that communication links between clients are bidirectional, and the matrix $G$ should be symmetric.

In order to find the collaboration graph, G, in a decentralized manner, we employ a greedy approach that involves identifying the $T$ closest similar models for each client. 
First, each client calculates its similarity with a fixed number of randomly selected other clients ($H$) using the model similarity metric in Equation \ref{eq:l1_similarity} ($H \ll T$).
Secondly, If two clients mutually select each other as their most similar clients among the $H$ samples, they form a two-member group. 
Clients not yet assigned to a group, for example, client $t$, can join the most similar client $k$, who has not already formed a group, without considering whether client $t$ is also the most similar among client $k$'s samples.
For the remaining clients without group affiliations, they randomly select another ungrouped client to create pairs, ensuring that all clients in the system become part of exclusive two-member groups.
Thirdly, Clients within the same group share their computed similarities with their group members. Each group then computes it similarity with other groups, approximated as the maximum similarity between one of its member and one member of the target group member.
The process then loops back to the second step, and these steps are reiterated until the desired group size is achieved.

Our goal is to group clients without introducing additional privacy risks. 
Therefore, we use clients' model weights after their first local training step. 
Since training is based on differential privacy (DP) guarantees, which we explain in the next section, sharing the output weights of the first iteration (used in group formation phase) does not compromise the privacy of individual clients. 
    
\para{Phase 2 Private co-training.} 
After group formation, clients within a group perform collaborative training.
To protect data privacy leakage, we need to apply DP noise during model training. However, in a fully decentralized system, it is challenging because the added DP noise could have a stronger negative effect on the accuracy of the training algorithm compared to centralized training. 
This is because the number of clients which share the knowledge is going to be smaller to match clients’ limited computation and communication power. Therefore, there is a need for a more robust feature against DP noise and an algorithm to aggregate knowledge between clients in an efficient way to make the learning process converge within a few iterations and end up with personalized parameters for each client.

In \name, we propose the use of transformed features from a handcrafted layer as the input to the client's neural network model. 
These handcrafted features can lead to better convergence, even in the presence of differential privacy noise \cite{tramer2020differentially}, and can achieve good performance even with a shallow neural network (as we demonstrate in Section~\ref{section:results}).
The transformed features are input into two models -- a private model, $f_{\theta_i}$ and a proxy model, $f_{w_i}$.
The client shares only the proxy model with other clients in its group. 
During each round of collaborative training, both the private and proxy models of a client $i$ are trained on local training data, $D_i$, with the difference being that the gradients of the proxy model are clipped and differential privacy noise is added. 
Given that the proxy model is trained using differential private training, sharing the model gradients $g_i = \dfrac{\partial w_i}{\partial x_i}$ with other clients does not violate DP guarantees.
Each client shares the computed gradients of its proxy model with selected neighboring clients and receives gradients from them to aggregate with its proxy model. 
By doing this, the proxy model $f_{w_i}$ incorporates information from the local data distribution of client $i$ and all the other members of client $i$'s group. 

To train the proxy model and private model together, inspired by \cite{kalra2023decentralized}, we apply knowledge distillation between these two models \cite{zhang2018deep}. More specifically, training of the private model will be done by classification loss (Equation~\ref{eq:proxy_classification}) and KL divergence loss (Equation~\ref{eq:proxy_kl}):

\begin{equation}
    \label{eq:proxy_classification}
    \mathbf{L}_{CE} (f_{w_i}) = \mathbb{E}_{(x,y) \sim D_i} CE[f_{w_i}(x) \parallel y]
\end{equation}

In Equation~\ref{eq:proxy_classification}, $CE$ represents the cross-entropy loss, which is calculated between the proxy model's output and the ground truth label $y$ corresponding to the input data $x$. 

\begin{equation}
    \label{eq:proxy_kl}
    \mathbf{L}_{KL} (f_{w_i} ; f_{\theta_i}) = \mathbb{E}_{(x,y) \sim D_i} KL[f_{w_i}(x) \parallel f_{\theta_i}(x)]
\end{equation}

In addition, to distill knowledge from the private model to the proxy model, we use the Kullback–Leibler divergence described in Equation~\ref{eq:proxy_kl} to make the proxy model output closer to the private model output.

Therefore, the proxy model learns from both the private model
and the local data by combining the classification loss and the KL divergence loss shown in equation \ref{eq:proxy_loss}. This way, the proxy model learns to correctly predict the true label of training instances as well as to match the probability estimate of its private model.

\begin{equation}
    \label{eq:proxy_loss}
    \mathbf{L}_{w_i} = (1- \alpha) \cdot \mathbf{L}_{CE} (f_{w_i}) + \alpha \cdot \mathbf{L}_{KL} (f_{w_i} ; f_{\theta_i}) 
\end{equation}

In the combination of the losses, $\alpha \in [0,1]$ balances between losses. As it increases, the proxy model will use more information from the private model. 

The objective of the private model is shown in Equation~\ref{eq:private_loss}.

\begin{equation}
    \label{eq:private_loss}
    \mathbf{L}_{\theta_i} = (1-\beta) \cdot \mathbf{L}_{CE} (f_{\theta_i}) + \beta \cdot \mathbf{L}_{KL}(f_{\theta_i} ; f_{w_i})
\end{equation}

As $\beta \in [0,1]$ increases, the private model will rely less on its local data and more on the data from other groups, which is in the proxy model. If $\beta$ is set to zero, it results in a totally personalized model that is only trained on the local training data.

With this architecture, the private model can extract personalized model weights for the client without dealing with DP noise, leading to improved performance. 
In other words, by decoupling private models from the differential privacy noise, we aim to enhance model performance, resulting in models that better align with the specific data characteristics of individual clients. Beside that, the parameters of proxy model could be aggregated with neighbor clients to benefit from their knowledge under DP guarantee.

 We note that it is not necessary for each member of a group to receive model updates from all other members and update its proxy model. 
 Instead, every few rounds, one client can volunteer within the group to act as an aggregator. 
 Therefore, clients within a group only send their model weights to the aggregator client and subsequently receives the aggregated model from that client, which is used for the next round of collaborative training.

\parait{Differential Privacy}

As clients share their gradients with other clients, we aim to control the information leakage from individual data of each client $D_i$ in the shared gradients. 
Similar to prior work~\cite{noble2022differentially}, we focus on record-level DP with respect to the joint dataset $D$ where $D$ and $D^{\prime}$ are neighboring datasets if they differ by, at most, one record. 
\cite{noble2022differentially} aims to quantify privacy level both towards a third-party observing the final model and honest-but-curious server.
As suggested by Noble \etal~\cite{noble2022differentially}, we set the DP budget in advance, denoted by $(\epsilon,\delta)$ and use the same  clipping and Gaussian noise to achieve this budget. 
Each gradient of client $i$, denoted by $g_i$, is divided by its norm as described in Equation~\ref{eq:clip} and then the noise is added to the gradient as shown in Equation~\ref{eq:noise}.

\begin{equation}
    \label{eq:clip}
    \Tilde{g_i} = \frac{g_i}{\frac{\max(1,\| g_i \|)}{C}}
\end{equation}

where $C > 0$ is the clipping parameter and $g_{ij}$ represents the gradient calculated on $i$-th client model on its $j-$th data.  

\begin{equation}
    \label{eq:noise}
    \Tilde{H_i} = \frac{1}{sR} \Sigma_{j\in S_i} \Tilde{g}_{ij} + \frac{2C}{sR} \mathcal{N}(0,\sigma_g^2)
\end{equation}
As showed in the equation, the std of Gaussian noise is $\sigma_g$ which will be calculated based on proposed bound of the paper which is shown in equation \ref{eq:dp_noise}.

\begin{equation}
    \label{eq:dp_noise}
    \sigma_g = \Omega( \frac{s \sqrt{lTK \log(\frac{2Tl}{\delta}) \log(\frac{2}{\delta})}} {\epsilon \sqrt{M^\prime}})
\end{equation}

As \cite{noble2022differentially} considered centralized FL, $l$ represents ratio of clients participate in each iteration of the federated learning. $K$ is the number of local steps which is conducted locally by clients and after the $K$ local steps, gradients will be shared. $M^\prime$ represents the number of gradients that will be aggregated. 
Since in P2P learning, individual gradients are directly shared among clients rather than being aggregated, we set $M^\prime$ and $l$ to 1.
In the paper, they proved that if the noise followed equation \ref{eq:dp_noise}, we could obtain ($\mathcal{O}(\epsilon_s),\delta_s$)-DP towards other clients (third-parties) which would receive the client's gradients in each communication round (after local training). 
We further mention how we set $\delta$ values based on a target $\epsilon$ in Section \ref{sec:evaluation}.

\section{Evaluation}
\label{sec:evaluation}
In this section, we analyze the trade-off between privacy and accuracy of \name and previous methods under different levels of data heterogeneity. 
Our analysis shows that our method is more robust against heterogeneous data and provides a better trade-off, leading to better performance over different model architectures (Linear neural network and CNN). We also study the practicality of deploying \name in resource-constrained environments. 

\subsection{Experimental setup}
\parait{Datasets.}
In our experiments, we consider image classification in distributed learning as the use case. 
To establish a distributed scenario, drawing inspiration from the non-IID setting employed in FedAvg \cite{mcmahan2017communication}, we randomly assign various classification tasks and their corresponding training data to each client. The generation of the non-IID tasks results in $M$ clients, where each client possesses $R$ samples, divided into 80\% for training and 20\% for testing. Among these $M$ clients, 20\% are reserved for evaluation purposes, specifically for hyper-parameter tuning. 

In order to generate non-IID tasks and explore the impact of varying data heterogeneity, we follow two common procedures used in previous works, which we refer to as i) sharding-based heterogeneity and ii) alpha-based heterogeneity. Both of these procedures partition the data samples among clients to achieve non-IID distribution. In the alpha-based method, the majority of data associated with a user originates from a single class. Conversely, in the sharding-based approach, each client's data comprises samples from specific multiple classes exclusively.
Specifically, sharding-based heterogeneity follows a procedure outlined by Li \etal~\cite{li2022learning}, where given a dataset consisting of $L$ classes, samples corresponding to each class are divided into $P$ shards. The objective is to generate $M = \frac{LP}{N}$ tasks, each defined by $N$ random classes with one random shard of data per class. To assess the impact of heterogeneity levels on different methods, we set $N$ to be in the set ${2,4,8}$.
Alpha-based heterogeneity is a data-generation procedure followed by prior work~\cite{noble2022differentially,hsu2019measuring}, where a  parameter $\gamma$ is set to define the heterogeneity of each client's data. For each client, $\gamma \%$ of the training and test data is sampled uniformly from all classes in an IID manner, while $1-\gamma \%$ is sampled from a specific class corresponding to the user.
To evaluate the impact of heterogeneity levels on different methods, we consider heterogeneity levels $\gamma$ in the set ${25\%,50\%,75\%}$.

We conduct experiments on three benchmark datasets, FEMNIST~\cite{cohen2017emnist}, CIFAR-10~\cite{krizhevsky2009learning} and CIFAR-100~\cite{krizhevsky2009learning}. 
CIFAR-10 and CIFAR100 consist of $32*32$ colored images, whereas FEMNIST consists of $28*28$ grayscale images. 
More statistics corresponding to datasets are showed in table \ref{tab:datasets}.

\begin{table}[]
    \caption{Non-IID tasks are randomly generated using alpha-based and shard-based generation procedures for the datasets. Each dataset consist of $L$ classes, which are partitioned among $M$ clients, with each client having $R$ samples in the local dataset.}
    \label{tab:datasets}
    \begin{tabular}{cccc} 
        \toprule
        Dataset   & L   & M   & R   \\ 
        \midrule
        FEMNIST    & 47  & 200 & 300 \\ 
        CIFAR-10  & 10  & 260 & 200 \\ 
        CIFAR-100 & 100 & 60  & 250 \\
        \bottomrule
    \end{tabular}
\end{table}

\parait{Model and Hyperparameters.}
To evaluate the effectiveness of our proposed methods and compare with related works, we conduct experiments using both a shallow neural network (linear layer with softmax activation function) and a CNN-based architecture~\cite{tramer2020differentially}. To ensure a fair comparison between the proposed methods, we conduct parameter tuning within a fixed range of parameters for all methods.
More specifically, following Noble \etal~\cite{noble2022differentially}, we set the global step-size to $\eta_g=1$ and the local step-size to $\eta_l=\frac{\eta_0}{sK}$, where $\eta_0$ is carefully tuned. Regarding privacy, we maintained a fixed value of $\delta=\frac{1}{R}$ in all experiments. For each setting, with the parameters related to sampling and the number of iterations fixed, we calculate the corresponding privacy bound $\varepsilon$ following \cite{noble2022differentially}.

\subsection{Implementation details}
\label{subsec:implementation_details}
As described in Section~\ref{section:design}, we use handcrafted features instead of raw images. 
Specifically, as proposed by Tramèr \etal~\cite{tramer2020differentially}, we use the scattering network of Oyallon \etal~\cite{oyallon2015deep} as a feature extractor that encodes images using wavelet transforms \cite{bruna2013invariant}. 
Following the default parameters outlined by Oyallon \etal~\cite{oyallon2015deep}, we utilize a ScatterNet $S(x)$ of depth two with wavelets rotated along eight angles. 
For an image of size $H * W$, the output dimension of the handcrafted feature extractor layer will be $(K,\frac{H}{4},\frac{W}{4})$, where $K$ equals 81 for grayscale images and 243 for RGB images. 
Additionally, we implement data normalization that normalizes each channel of $S(x)$. 
Each client calculates the mean and variance of its local data, thereby incurring no additional privacy cost for the normalization step.
Utilizing these transformed features instead of raw images offers several advantages, including improved accuracy performance, and the need for only one linear layer to classify images, as we will show in Section~\ref{section:results}.

\subsubsection{Baselines}
For comparison between our proposed method and previous works, we study the performance of different baseline algorithms under varying levels of heterogeneity. 
We use the following baselines:
\begin{itemize}
    \item \textit{Centralized learning.} In centralized learning, we consider that all the data is stored in a centralized location (without concern for the privacy of different clients), and that there is a server with high computational power to train a model with all the training datasets. This scenario is considered to be an optimal scenario and provides a measure of the classification difficulty. We report the results of the centralized model with and without handcrafted features.

    \item \textit{Local training.} In local training, each client exclusively trains its model with its local dataset. Initially, this may appear to be a weaker baseline. However, our findings reveal that in cases of high data heterogeneity, knowledge sharing exhibits only modest improvements compared to local training. Notably, in such scenarios, most previous approaches fall short of achieving the same accuracy as local training.
    
    \item \textit{FedAvg.} FedAvg is a widely-used federated learning algorithm~\cite{mcmahan2017communication}. In FedAvg, there is a centralized server with high computational power and all clients have a communication link to the server. In this baseline, we assume that the server is not completely trustworthy and is honest but curious. Therefore, we consider the corresponding privacy bound $\epsilon$ with respect to the server.

    \item \textit{Scaffold}. As a centralized federated algorithm, Scaffold~\cite{noble2022differentially} addresses data heterogeneity under DP constraints. By introducing an additional parameter for each client, it aims to consider the drift shift of the client from the server model to achieve better personalization in FL compared to FedAvg. Similar to FedAvg, we assume that the server is untrustworthy, so the calculation of the privacy bound is with respect to the server. Privacy loss is calculated based on their proposed subsampled Gaussian mechanism under Renyi Differential Privacy (RDP)~\cite{mironov2017renyi}.

    \item \textit{ProxyFL.} ProxyFL is one of the state-of-the-art methods in decentralized learning. In ProxyFL, clients have a private model which they do not share, and a proxy model which they share with other client. We assume that the proxy model and private model have the same architecture. We also follow the same communication method (directed exponential graph) and method for differential privacy as described in the ProxyFL paper~\cite{kalra2023decentralized}.

    \item \textit{DP-DSGT}: In \cite{bayrooti2023differentially}, three methods were proposed to provide DP in a P2P learning scenario. Among them, DP-DSGT claims to have consistent performance regardless of data distribution, making it suitable for applications where agents have non-overlapping data classes. This paper represents one of the state-of-the-art approaches in differentially private P2P learning.
\end{itemize}

Although some previous works have explored clustering clients in decentralized learning \cite{li2022towards,duan2021flexible,xie2021multi,sattler2020clustered,nguyen2022self,briggs2020federated,ghosh2020efficient}, all of these methods either require a centralized server to perform clustering between clients or violate differential privacy. Therefore, we exclude these methods from our performance comparison with \name.

\subsection{Results}
\label{section:results}
\begin{figure*}[h]
    \centering
    \subfigure[]{\includegraphics[width=0.33\textwidth]{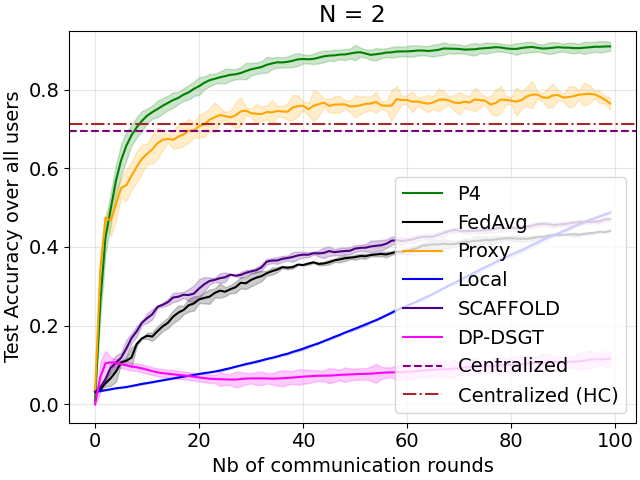}}
    \subfigure[]{\includegraphics[width=0.33\textwidth]{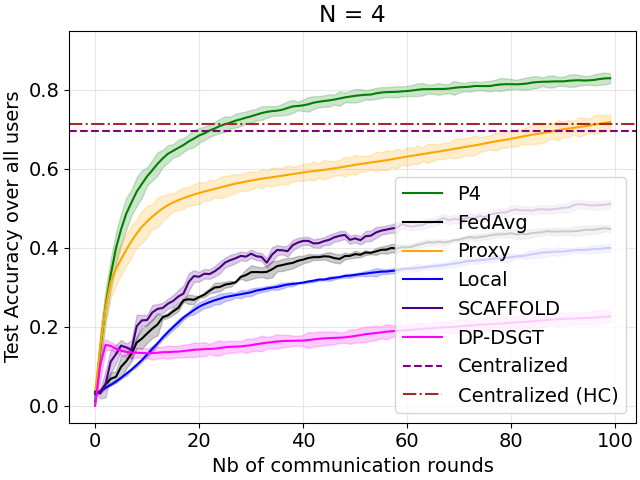}} 
    \subfigure[]{\includegraphics[width=0.33\textwidth]{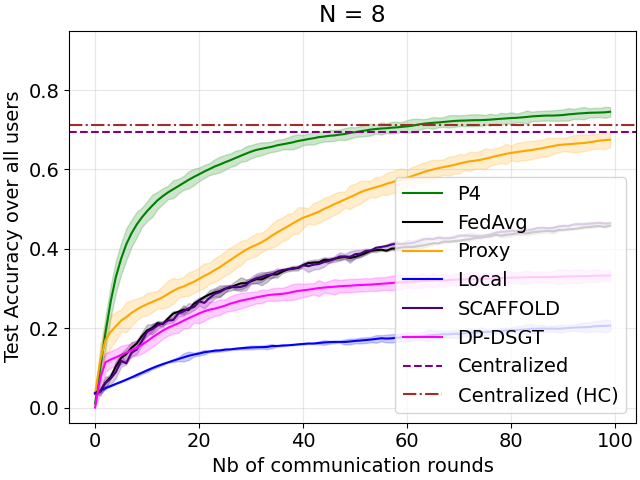}}
    \caption{Test Accuracy for Linear model architecture on FEMNIST with $\epsilon=15$ (a) $N=2$ (b) $N=4$ (c) $N=8$}
    \label{fig:shard_FEMNIST_NN2}
\end{figure*}

\begin{figure*}[h!]
    \centering
    \subfigure[]{\includegraphics[width=0.33\textwidth]{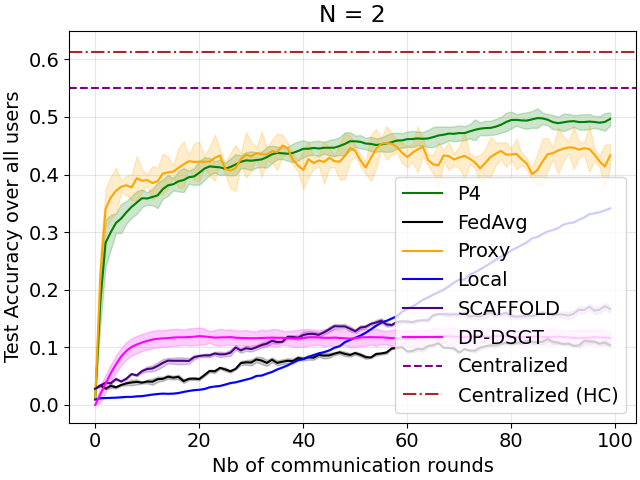}}
    \subfigure[]{\includegraphics[width=0.33\textwidth]{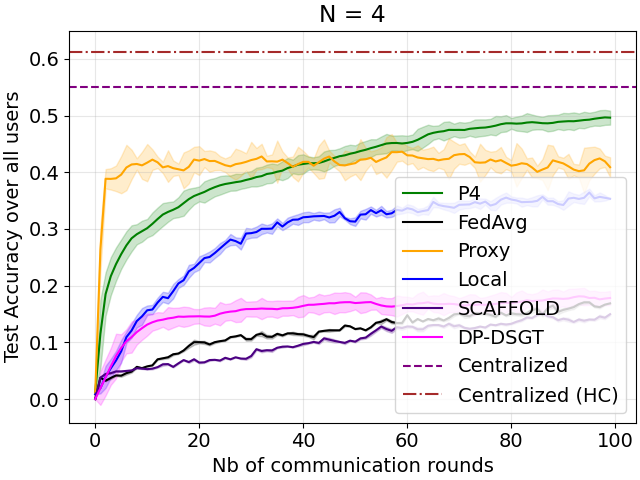}} 
    \subfigure[]{\includegraphics[width=0.33\textwidth]{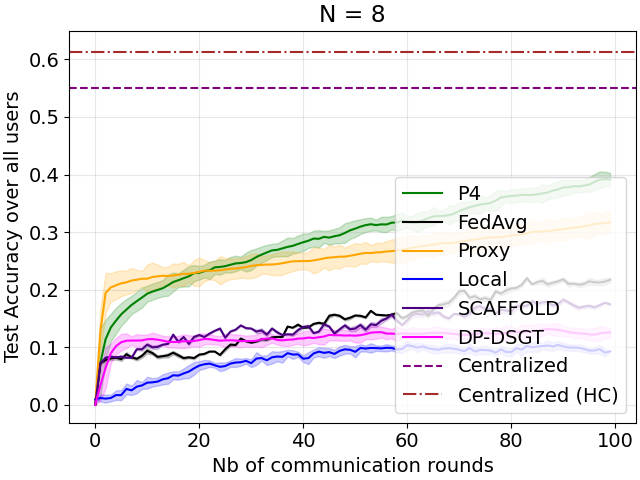}}
    \caption{Test Accuracy for Linear model architecture on CIFAR-100 with $\epsilon=15$ (a) $N=2$ (b) $N=4$ (c) $N=8$ }
    \label{fig:shard_CIFAR100_NN2}
\end{figure*}

\begin{figure*}[h]
    \centering
    \subfigure[]{\includegraphics[width=0.33\textwidth]{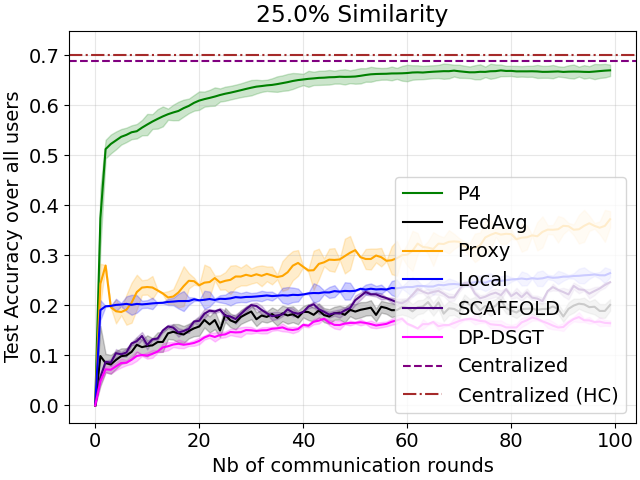}}
    \subfigure[]{\includegraphics[width=0.33\textwidth]{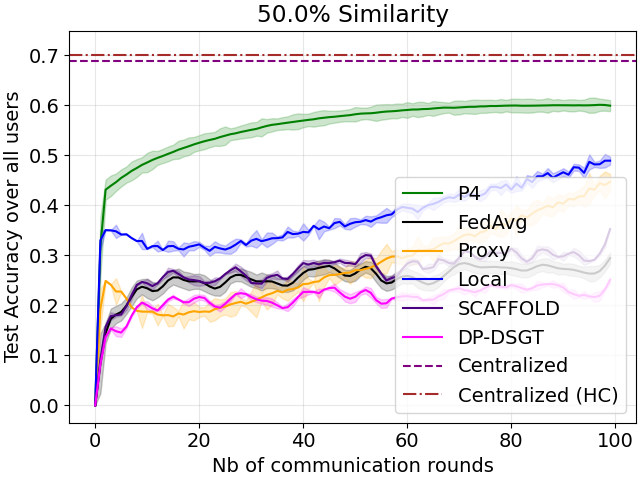}} 
    \subfigure[]{\includegraphics[width=0.33\textwidth]{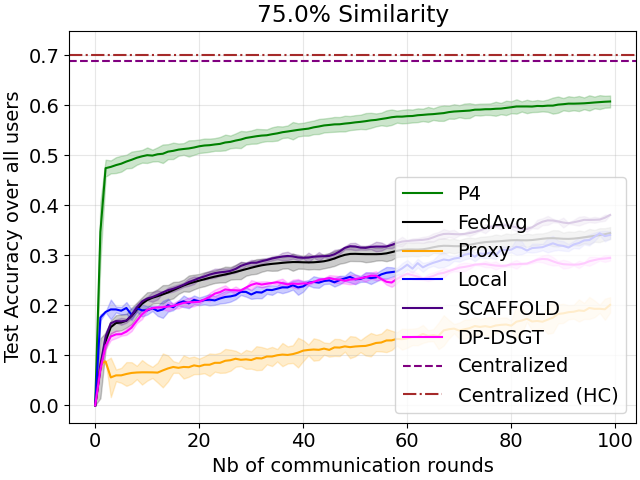}}
    \caption{Test Accuracy for CNN model architecture on CIFAR-10 with $\epsilon=15$ (a) $\gamma=25\%$ (b) $\gamma=50\%$ (c) $\gamma=75\%$ }
    \label{fig:diff_sim_cnn}
\end{figure*}

\begin{figure*}[h]
    \centering
    \subfigure[]{\includegraphics[width=0.33\textwidth]{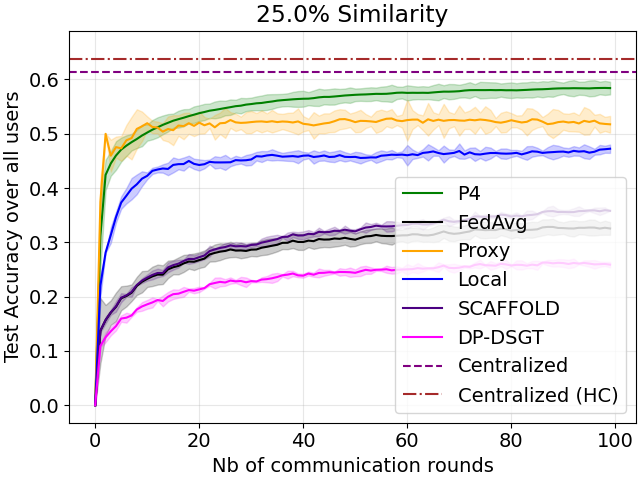}}
    \subfigure[]{\includegraphics[width=0.33\textwidth]{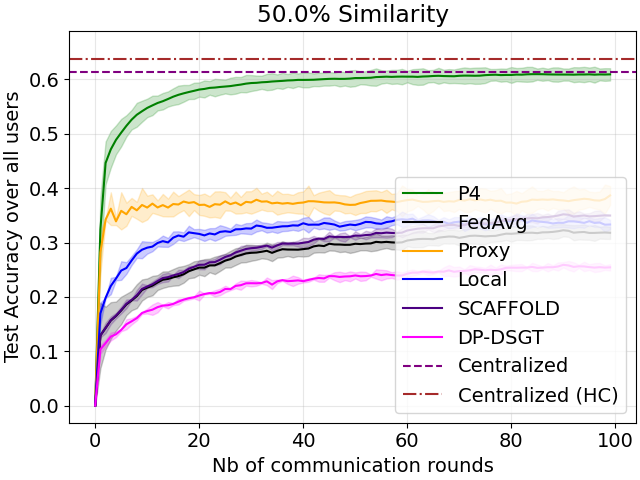}} 
    \subfigure[]{\includegraphics[width=0.33\textwidth]{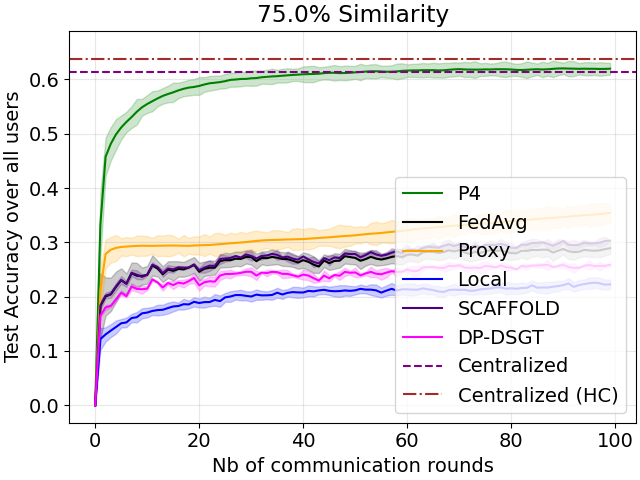}}
    \caption{Test Accuracy for Linear model architecture on CIFAR-10 with $\epsilon=15$ (a) $\gamma=25\%$ (b) $\gamma=50\%$ (c) $\gamma=75\%$ }
    \label{fig:diff_sim}
\end{figure*}

\begin{figure*}[h]
    \centering
    \subfigure[]{\includegraphics[width=0.33\textwidth]{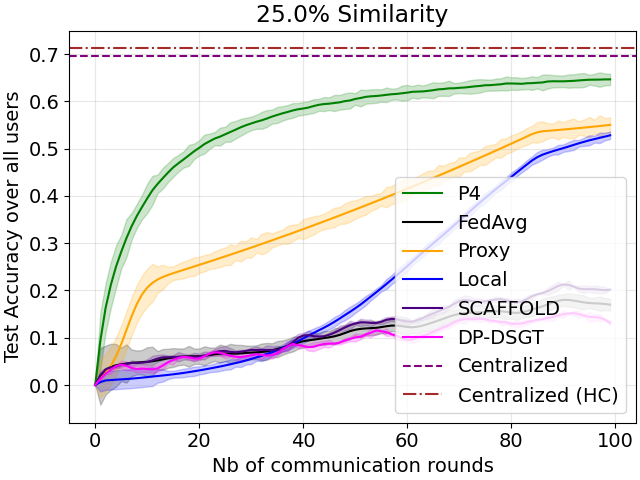}}
    \subfigure[]{\includegraphics[width=0.33\textwidth]{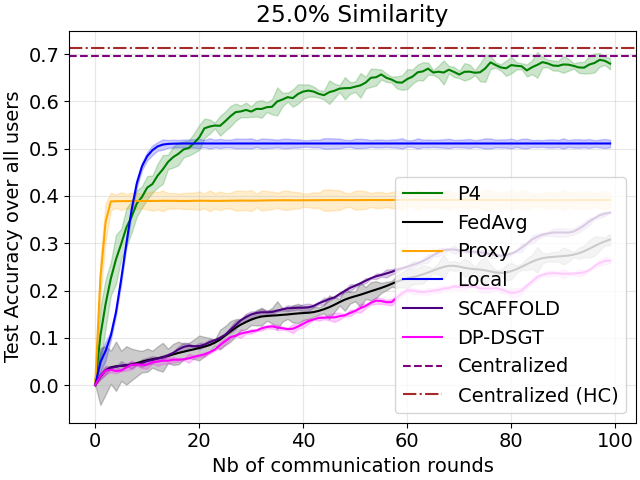}} 
    \subfigure[]{\includegraphics[width=0.33\textwidth]{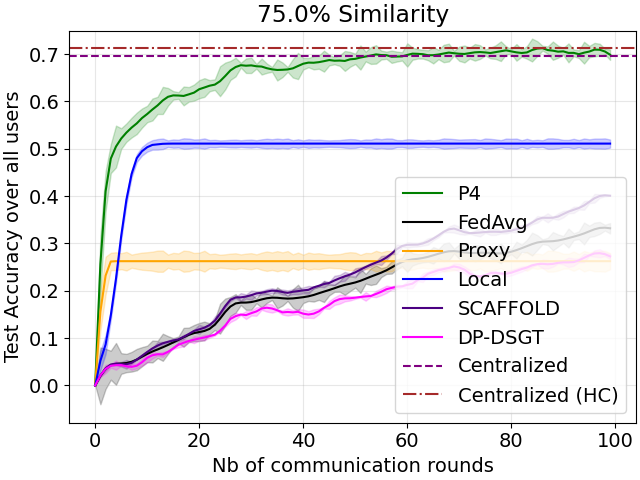}}
    \caption{Test Accuracy for Linear model architecture on FEMNIST with $\epsilon=15$ (a) $\gamma=25\%$ (b) $\gamma=50\%$ (c) $\gamma=75\%$ }
    \label{fig:emnist_results}
\end{figure*}

Firstly, we analyze the performance of various algorithms under different types and levels of data heterogeneity by varying $\gamma$ and $N$, corresponding to alpha-based and shard-based data heterogeneity, respectively.
In this regard, we fixed the number of training communication rounds $T = 100$ to mimic a scenario in which each client has limited computation power and the goal is to converge to a good performance on its data distribution fast (with a limited number of iterations). 
Regarding differential privacy, we fixed target guarantee $\epsilon=15$ for all methods. 
Then, we consider all combination hyper-parameters: local steps $K\in {1,2,...,10}$ , user sampling ratio $u \in {0.1,0.2,...,1}$ and data sampling ratio $s\in{0.1,0.2,...,1}$ and calculate the corresponding noise level $\sigma_g$ using Equation~\ref{eq:dp_noise} to achieve the target epsilon. 
Furthermore, we set the group size for our method to $|g|=4$ for experiments on CIFAR-100 and $|g|=8$ for the remaining experiments.
To have a fair comparison between different methods, we perform hyper tuning (grid search) on the mentioned hyperparameters and the local step size $\eta_l \in [0.1,10]$, clipping norm $C \in [0.1,10]$. We conduct hyper tuning  on the evaluation data which is mentioned in Section~\ref{subsec:implementation_details}.
The performance of different methods under shard-based data heterogeneity is depicted in Figures~\ref{fig:shard_FEMNIST_NN2} and~\ref{fig:shard_CIFAR100_NN2}, utilizing the linear model on FEMNIST and CIFAR-100 datasets, respectively.
Moreover, the methods' performance under varying levels of heterogeneity of alpha-based procedure are illustrated in Figure~\ref{fig:diff_sim},~\ref{fig:emnist_results} (for the linear model) and Figure~\ref{fig:diff_sim_cnn} (for the CNN model).
In the figures, ``(HC)" indicates whether handcrafted features are used.
We conduct each experiment 3 times to reduce the randomness effect and have more reliable results, so we report the mean of the results across trials in the following figures.

As illustrated in the figures, our approach consistently outperforms previous proposed methods across varying levels and types of data heterogeneity. Our method achieves an accuracy ranging from $58.6\%$ to $62.2\%$ on CIFAR-10 based on alpha-based data generation, utilizing just a single linear layer, making it highly suitable for resource-constrained devices. In contrast, FedAvg struggles to achieve satisfactory performance in the presence of data heterogeneity, highlighting the inherent problem of centralized federated learning with non-IID data distribution of clients' data. Scaffold as other centralized method performs better than FedAvg specially when the level of data heterogeniouty is significant (small N and small $\gamma$). Although it exhibits capability in personalized federated learning, it still struggles to converge rapidly to a proper solution, as demonstrated in both data generation methods.

As shown in Figure~\ref{fig:shard_FEMNIST_NN2}, DP-DSGT, as a P2P method, achieves performance below the centralized baselines since its upper-bound performance is limited to achieving consensus among all clients (output of centralized methods). 

By considering the performance of FedAvg, Scaffold, and DP-DSGT on different datasets and architectures, we can conclude that when the tasks of clients are non-IID, consensus learning may not achieve good performance, as shown in the experiments.

In our experiments with a CNN model in Figure~\ref{fig:diff_sim_cnn}, which possesses more parameters compared to the linear model, previous approaches suffered from noisy training due to differential privacy noise and collaborative training with clients with different data distribution. In contrast, \name consistently converges to robust parameter solutions across all settings, even with a minimal number of communication rounds.

Our proposed model consistently delivers mostly stable performance across various levels of data heterogeneity. Unlike other methods that excel at specific levels of heterogeneity while struggling in other scenarios (e.g., In Figure \ref{fig:emnist_results}, the proxy paper at $\gamma = 25\%$ exhibits good performance; however, in certain scenarios, it does not yield a significant improvement over local training.), our approach consistently outperforms others in both architectures across different heterogeneity levels.

Figures~\ref{fig:shard_FEMNIST_NN2},~\ref{fig:shard_CIFAR100_NN2}, ~\ref{fig:diff_sim} and~\ref{fig:emnist_results} show that when a client's local dataset has just a few classes and uses a simple neural network, \name can perform similarly or even better than centralized training in some cases. This is because a shallow neural network might find it easier to learn the patterns needed to classify a limited portion of the data distribution, compared to the complexity of classifying samples from all classes (whole data distribution). In other words, due to the limited number of model parameters in these scenarios, classifying the entire data distribution poses a challenge for the centralized method. Conversely, a personalized model for each client and implicitly for each data distribution would be even better than a centralized model.

Furthermore, our results indicate that local training can serve as a potent baseline for decentralized learning algorithms when data similarity among clients is limited. Consequently, addressing the challenge of effective collaborative training in highly heterogeneous data distributions becomes increasingly important to improve accuracy compared to local training.

\subsubsection{Comparison between collaborative training and local training}
\label{section:tradeoffDP}
\begin{figure}
    \centering
    \includegraphics[width=1\linewidth]{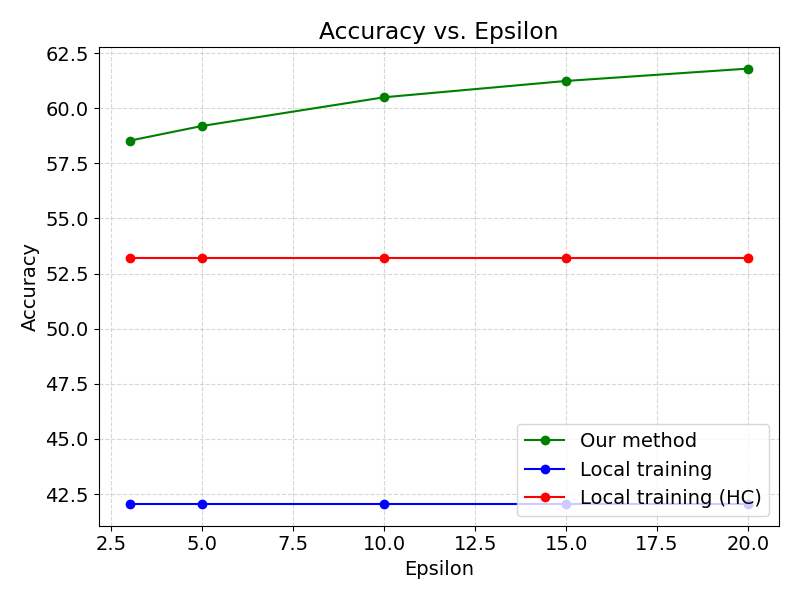}
    \caption{Performance of the linear model (on CIFAR-10) under different level of privacy boundaries and compared to local training.}
    \label{fig:tradeoff}
\end{figure}

Distributed learning involves more communication and privacy concerns compared to local training. 
In local training, clients improve their models using only their own data, which means that they do not need to communicate with others during training. 
To see how well our differential private collaborative training method performs compared to the non-DP version of local training and to understand the impact on privacy, we tested our method under different privacy settings, with privacy bounds ranging from $\epsilon = 3$ to $\epsilon = 20$. The tasks are generated using the alpha-based method, and we utilized the FEMNIST dataset for this study.
We compared this to how well local training performs on the previously mentioned linear model and data heterogeneity $\gamma = 50\%$.
Our findings, illustrated in Figure \ref{fig:tradeoff}, reveal that collaborative training outperforms local training even when the privacy budget is set higher than $3$. 
Even when strong privacy constraints are in place (e.g., $\epsilon = 3$), collaborative training performs better, addressing the issue of over-fitting that can affect local training due to its limited amount of local training data. Moreover, as shown in the figure, even though using handcrafted features could improve local training accuracy, it is still not as good as our proposed method. On the other hand, our method shows good robustness against DP noise such that it could have reasonable accuracy with restricted privacy.

\subsection{Ablation study}
\label{sec:ablation}
\begin{figure}
    \centering
    \includegraphics[width=1\linewidth]{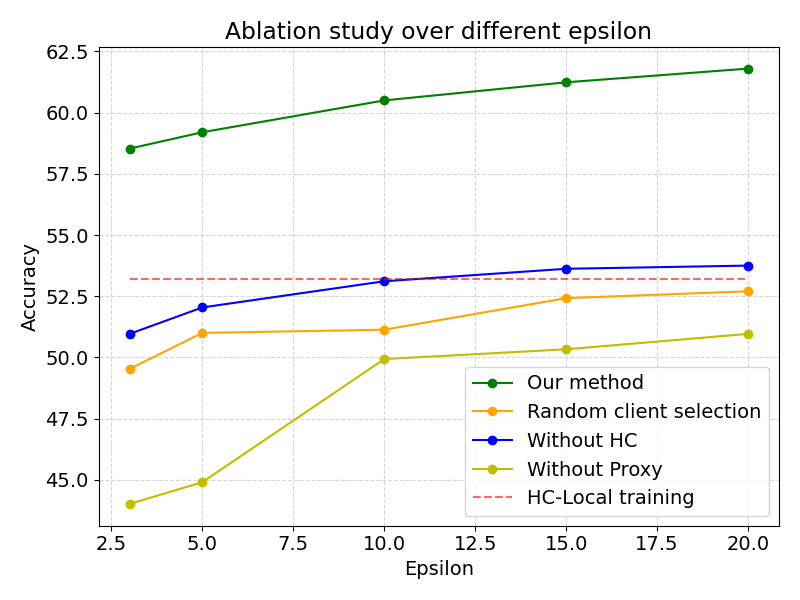}
    \caption{Comparing the effect of each component of our proposed method using the linear architecture and data similarity of $\gamma = 50\%$ (on CIFAR-10).}
    \label{fig:ablation}
\end{figure}

We perform an ablation experiment on CIFAR-10 using the linear model, aimed at understanding the effect of client selection, handcrafted features, and proxy model individually on the performance of \name. 
Therefore, we compare \name's accuracy results with three different methods: i) Random client selection instead of using our group clustering technique. ii) Using raw images instead of handcrafted features. iii) Removing the proxy model and using one model per client instead. 
In each experiment, one of these components is removed and the performance of the model is shown in Figure \ref{fig:ablation}. The ablation is 
As shown in the figure, each of these three components has a strong effect on the model performance -- removing even one of them results in accuracy lower than the local training baseline in most cases. 
Our study shows that we can achieve private personalized learning only with a well-designed method that considers all phases of communication among clients. 

\subsection{\name on resource-constrained devices}
\label{subsetion:energy}

We evaluate the practicality and overhead of \name by implementing it on resource-constrained devices.
We implement \name on two Raspberry Pis acting as clients in the system and evaluate its performance across four metrics: run time, memory usage, power consumption, and communication bandwidth. 

We use the Raspberry Pi Model 4B with the following settings: OS Debian 12 (Bookworm) 64-bit, Python version 3.11.2, PyTorch version 2.1.0. 
The clients have the Linear model on CIFAR-10 dataset  and communicate with each other over secure websockets.
We obtain the performance metric values for each phase of \name.
We measure the run time of \name using Python's \texttt{time} library, and the memory consumption using the \texttt{free} command.
In order to measure the power consumption, we plug in the Raspberry Pis to smart plugs (Tapo P110) and obtain the consumed power values using the Tapo API. 
For communication overhead, we measure the size of the transmitted data between the two clients. 
We provide the measurements for each metric below.

\textit{Run time.} We measure the average run times across 100 iterations each of both phases. We find that phase 1 (group formation) between two clients takes 0.04 seconds on average (std. 0.02). 
Assuming the same scenario as the experiment in Section \ref{section:results} (Each client samples 35 clients to compute its model similarity), it would take around 1.4 seconds in total to run phase 1.
Phase two (private co-training) between two clients takes an average of 5.27 seconds (std. 0.58) to complete, with the bulk of the run time being the training process (avg. 4.83 seconds, std. 0.05). 

\textit{Memory usage.} We find that running \name consumes around 72 MB during the first phase and 489 MB memory during the second phase.  

\textit{Power consumption.} The baseline consumption of the Raspberry Pi is $\approx$ 2.64 W (std 0.01). We find that the average consumption during the first phase is 3.17 W (std. 0.31) and second phase is 4.87 W (std. 0.34). 

\textit{Communication bandwidth.} During the first phase, the client that initiates the communication sends its model weights to another client, which then performs the comparison with its own weights. We use the Python \texttt{pickle} library for message serialization. The message size of the weights is 622.82 kB. During the second phase, the client that initiates the co-training first sends its model parameters. After training, the recipient client sends back its gradients to the initiator client for aggregation. The total size of messages exchanged during this phase is 1246.57 kB. 

Our experiments indicate that \name can be run on real resource-constrained devices with minimal overhead.

\section{Conclusion}
\label{sec:conclusion}
We introduced a novel private personalized peer-to-peer learning approach, named \textit{\name}. 
Our method, operating in a fully decentralized manner, introduces a lightweight method to cluster clients based on model similarity that outperforms random clustering techniques. We develop a knowledge-sharing method among clients within a group that outperforms previous works over different levels of clients' data heterogeneity. A notable advantage of \name is its ability to be run on resource-constrained device because it converges to good accuracy on the benchmark datasets by using just a shallow neural network. We demonstrated the practicality of our method by implementing it on a Raspberry Pi and measuring its overhead in terms of power consumption, run time, memory usage, and communication bandwidth. 

Our work can be extended in a few ways. 
We used a similarity metric to group clients and conducted knowledge sharing within each group. 
We employed a simple approach to make groups based on limited sampling. 
Although this approach leads to good performance, optimizing the algorithm that makes groups based on sampled similarities could be a potential avenue for improvement. 
The optimization would be in terms of reducing communication overhead or grouping clients more precisely such that sampling has minor negative impact on the accuracy of the overall training algorithm. 

Additionally, we considered a setting with honest but curious clients. 
Improving \name to be robust against malicious clients that can share incorrect or harmful inputs that affect the training algorithm is another area for future work.

Finally, handcrafted features are crucial in our design, enabling our algorithm to converge even with noisy gradients. 
While we showed the benefit of handcrafted features in image classification, the use of handcrafted features could potentially be extended to other domains, such as text and audio, by employing techniques proposed in previous works \cite{manning1999foundations} and \cite{anden2014deep}, respectively.

\bibliographystyle{ACM-Reference-Format}
\bibliography{paper/bibliography}

\appendix

\clearpage
\onecolumn

\end{document}